\documentclass[11pt]{article}
\PassOptionsToPackage{hyphens}{url}
\usepackage[final]{acl}
\usepackage{times}
\usepackage{latexsym}
\usepackage[T1]{fontenc}
\usepackage[utf8]{inputenc}
\usepackage{microtype}
\usepackage{graphicx}
\usepackage{booktabs}
\usepackage{amsmath,amssymb}
\usepackage{tikz}
\usetikzlibrary{arrows.meta,positioning}

\title{Measuring the Serving Stack Instead of the Model:\\
       Hidden Confounds in Local Tool-Use Evaluation}

\author{Lijuan Tang \\ Northeastern University, Seattle \\ \texttt{tang.lij@northeastern.edu}
  \And Yuemeng Zheng \\ Northeastern University, Seattle \\ \texttt{zheng.yuem@northeastern.edu}}

\begin{document}
\maketitle

\begin{abstract}
A coding agent must emit a valid tool call---a parseable invocation of a tool in the
provided schema---before the harness can execute its chosen action. We study how local
serving stacks affect this protocol step and show that measured outcomes can depend on
the serving layer rather than model behavior alone. In Ollama, the default
\texttt{tools=} request is gated per model by a static template flag: some models are
accepted and return calls as text, some return native \texttt{tool\_calls}, while
Phi-3 and Gemma-3 are rejected before inference. In our harness, rejection and retry
exhaustion are not preserved as structured failure metadata, so downstream analysis
can misclassify them as model non-calls and naively report $0\%$ fidelity. Adding a text tool list while retaining the
native channel recovers much of the measured fidelity for accepted models, whereas a
uniform text protocol reduces fidelity for Llama-3.2, which has native tool-call
support. Cross-stack probes on Ollama, llama.cpp, vLLM, and SGLang show different
handling of the same request. Constrained decoding removes parse failures but can
induce non-termination, and turn-pooled versus per-instance estimates differ by up to
about 55 points. We conclude with a checklist for treating serving behavior as part of
the evaluation protocol.
\end{abstract}

\section{Introduction}
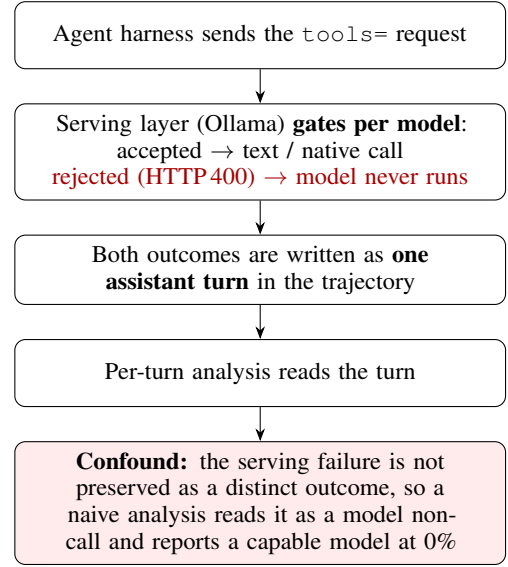
\begin{figure}[t]
\centering
\begin{tikzpicture}[
  node distance=4.5mm,
  box/.style={draw,rounded corners,align=center,font=\small,
              text width=6.2cm,inner sep=4pt,minimum height=2.3em},
  >=Stealth]
\node[box] (h) {Agent harness sends the \texttt{tools=} request};
\node[box,below=of h] (s) {Serving layer (Ollama) \textbf{gates per model}:\\
  accepted $\to$ text / native call\\
  \textcolor{red!65!black}{rejected (HTTP\,400) $\to$ model never runs}};
\node[box,below=of s] (t) {Both outcomes are written as \textbf{one assistant turn}
  in the trajectory};
\node[box,below=of t] (m) {Per-turn analysis reads the turn};
\node[box,below=of m,fill=red!8] (c) {\textbf{Confound:} the serving failure is not
  preserved as a distinct outcome, so a naive analysis reads it as a model
  non-call and reports a capable model at 0\%};
\draw[->] (h)--(s); \draw[->] (s)--(t); \draw[->] (t)--(m); \draw[->] (m)--(c);
\end{tikzpicture}
\caption{How the serving layer silently contaminates the measurement. The harness
records a refused request as an ordinary non-call turn: the transport error is
received but not persisted as structured metadata, so a downstream per-turn analysis
reads it as a model that declined to call a tool unless it string-matches the
harness's error text (\S\ref{sec:audit}).}
\label{fig:pipeline}
\end{figure}

A coding-agent loop assumes the model can reliably emit a tool call the harness can
execute. This is a protocol step, separate from the semantic step of choosing the
right tool. We wanted to measure, on small local models, how often this step
succeeds. The measurement turned out to be the problem (Fig.~\ref{fig:pipeline}).
Our contribution is therefore not another tool-use benchmark but a measurement study:
it shows that serving-layer behavior can systematically confound the evaluation of
local coding agents, so a standard setup can measure the stack rather than the model.

Every local agent stack interposes a serving layer between the harness and the model
weights, and that layer decides how a tool-call request is handled before the model is
ever consulted. We use Ollama as the primary case study and llama.cpp as a cross-check,
but the mechanism is a property of the serving layer, not of either library. With
models served through Ollama's OpenAI-compatible endpoint, the default harness sends the
\texttt{tools=} parameter, and Ollama gates that request per model using a static flag
in the model template rather than the model's generated behavior. On one
fixed stack we see three behaviors. Qwen2.5-Coder (all sizes) accepts \texttt{tools=}
and returns the call as text. Llama-3.2 returns native \texttt{tool\_calls}. Phi-3
and Gemma-3 have the request rejected with HTTP~400 ``does not support tools''. The
harness retries, gives up, and writes a generic error string into the trajectory as
an assistant turn (\S\ref{sec:audit}); the error type is not preserved, so a
downstream per-turn analysis reads it as a model that chose not to call a tool. A
capable model is then reported at $0\%$ because its request never ran, and the
failure is silent.

We summarize the serving-layer mechanics this depends on in \S\ref{sec:method}.

This is not a claim that small models are good or bad at tool use. It is a validity
threat that is easy to miss, because the contaminated number looks like an ordinary
result. We make the point precise with a set of experiments and one negative result:
\begin{itemize}
  \item \textbf{Native function calling (FC)}---the path in which the harness passes
  tool schemas in the request's \texttt{tools=} field for the server to route to the
  model, rather than describing them in the prompt text---\textbf{silently
  contaminates the measurement} through request rejection (Phi-3, Gemma-3 reported at
  $0\%$ without running) and through model non-responses, both logged as ordinary
  non-call turns.
  \item \textbf{A control arm separates the channel from the prompt.} By ``channel''
  we mean whether tool schemas travel in the \texttt{tools=} field or as prompt text.
  For models the
  server accepts, keeping the native channel but adding a text tool list and call
  format to the prompt recovers most of the fidelity; for Llama-3.2, the uniform
  text protocol lowers it. The best configuration is model-dependent.
  \item \textbf{A four-stack check} shows the same weights that Ollama rejects run
  on llama.cpp, while vLLM refuses the same request by default and SGLang accepts it
  but returns the call as text: four stacks handle the identical request differently,
  so the outcome is a serving-stack property, not the model.
  \item \textbf{The fidelity number is not robust.} Turn-pooled and per-seed rates
  differ by up to about 55 points because one looping episode dominates the pool,
  on thin denominators.
  \item \textbf{A constrained-decoding baseline} makes every model emit a valid call
  by construction, but makes the weaker models loop without terminating, which moves
  the failure instead of removing it.
  \item \textbf{Negative result.} We could not support claims about scale, family,
  or reasoning at this measurement precision, and we say so explicitly instead of
  asserting them.
\end{itemize}
The contribution is a measurement caveat and a checklist, not a leaderboard.

\section{Related Work}
\paragraph{Function-calling evaluation.} BFCL \cite{bfcl2025}, from the Gorilla line
of work connecting LLMs to external APIs \cite{gorilla2023}, is the standard
function-calling leaderboard and distinguishes native FC mode from prompting mode,
where tools are supplied in the prompt; our text-tools condition is an instance of
prompting mode, and our control arm sits between the two. BFCL can serve models
locally, retains raw responses, and already reports empty-turn and API-error
categories, so separating infrastructure outcomes from model outcomes is not itself
new. What it does not do is treat the serving interface as an experimental variable:
it evaluates through per-model handlers and reports aggregate accuracy rather than a
per-turn outcome composition on a held-fixed stack, so per-model request gating
inside a local serving layer is not isolated. Our contribution is that controlled
isolation across serving stacks, not the first separation of infrastructure outcomes. Agent and tool-use benchmarks such as
AgentBench \cite{agentbench2024}, $\tau$-bench \cite{taubench2024}, and ToolSandbox
\cite{toolsandbox2024} evaluate richer multi-turn and stateful interaction, and
$\tau$-bench likewise finds tool-use behavior unreliable across repeated trials; but
they target hosted or fixed model endpoints and do not isolate the local serving
stack as a source of measurement error. Coding-agent harnesses such as SWE-agent
\cite{sweagent2024} standardize the agent-computer interface and do persist
structured per-step records; the conflation we document is a property of the
particular ReAct harness we instrument, which retains only the assistant message
stream (\S\ref{sec:audit}), and we make no claim about harnesses we did not run.
\paragraph{Serving backends and harnesses as measurement variables.} Closest to our
setting, \citet{silenthyper2026} quantify how the choice of inference backend alone
changes LLM outputs and undermines reproducibility, and Harness-Bench
\cite{harnessbench2026} measures how the agent harness, with the model held fixed,
shifts outcomes in realistic agent workflows. Both establish infrastructure as a
first-class source of variance. Neither examines the per-model \texttt{tools=} gate
inside a local serving layer, nor the recording of a refused request as a model
non-call, which is the specific mechanism we isolate.
\paragraph{Constrained decoding as the local remedy.} The standard fix for
unparseable tool calls on local models is grammar-constrained or schema-guided
decoding (e.g.\ GBNF grammars, Outlines \cite{outlines2023}, vLLM guided decoding,
Ollama structured output), which constrains the format at the token level;
\citet{tooldec2023} apply finite-state decoding specifically to tool calls to
eliminate syntax errors by construction. We do not
propose a fix; we include constrained decoding only as a baseline
(\S\ref{sec:results}) to show the protocol failures are removable by construction,
and we measure how the unconstrained default setups mismeasure the model.
\paragraph{Tool hallucination and failure taxonomies.} PA-Tool \cite{patool2025}
names ``schema misalignment'', the hallucination of absent tool names, which we
observe as one outcome category. MAST \cite{mast2025} catalogs agent failures on
capable models and does not address the local serving layer. Across these lines of
work, existing benchmarks implicitly assume the serving layer is transparent; our
study questions that assumption.

\section{Method}
\label{sec:method}
\paragraph{Background: the \texttt{tools=} request and the serving layer.} In the
OpenAI-compatible protocol that local servers expose, a tool-call request carries a
\texttt{tools} array of function specifications---each a name, a description, and a
JSON-Schema parameter object---alongside the messages. The model never receives that
array directly. The serving layer must first render the specifications into the prompt
using the model's chat template, and then parse the generated text back into a
structured \texttt{tool\_call}. Both steps happen outside the weights, so whether a
server performs them is determined by the model's template and the server's launch
configuration rather than by what the model can do. Ollama exposes this per model as a
\texttt{capabilities} list through \texttt{/api/show}, and documents \texttt{tools} as
usable only ``if supported''; vLLM and SGLang instead require an explicit
\texttt{-{}-tool-call-parser} before an accepted call is recognized as one. The same
request and the same weights can therefore yield different recorded outcomes on
different stacks (Fig.~\ref{fig:threestacks}).
\paragraph{Harness and per-turn taxonomy.} We use an off-the-shelf ReAct
coding-agent harness, an extension of LOCA-bench \cite{locabench2026}, on a fixed
aggregation task that requires several tool calls. We label every assistant turn:
\textbf{valid in-schema call}, \textbf{hallucinated call} (parseable, tool not
found), \textbf{unparseable text}, \textbf{no-call prose}, and \textbf{non-response}
(the harness wrote its retry-exhaustion error, \S\ref{sec:audit}). Protocol fidelity
is the valid-in-schema-call rate over turns the model actually produced, so
non-responses are excluded from the denominator and reported separately. We run 8
seeds per model and report both the turn-pooled rate and the per-seed mean (each
episode weighted equally) with a seed-level bootstrap $95\%$ CI ($10{,}000$ resamples
of the per-seed rates); proportions on 4--8 seeds are non-normal, so the bootstrap is
more honest than a normal $\pm$SE interval. We write ``seed'' throughout for a task
instance (the harness's episode index), not a decoder RNG seed: decoding is sampled at
$T{=}1.0$ and is not held fixed, so variation across seeds mixes task and sampling
variation (\S\ref{sec:repro}).
\paragraph{Three serving conditions.} (i) \emph{native}: harness default, sends
\texttt{tools=}; on Ollama this is gated per model. (ii) \emph{native+hint}:
\texttt{tools=} still sent, plus a plain-text tool list and an explicit JSON call
format with an allowed-name list injected into the prompt; this isolates the prompt
guidance from the serving channel and is only defined for models the server accepts.
The hint is a single fixed rendering used for every model, not tuned per model: the
tool schemas as prose plus a strict JSON call format, mirroring BFCL's prompting mode
\cite{bfcl2025}. The explicit allowed-name list targets a known, model-agnostic
failure mode, tool-name hallucination \cite{patool2025}, rather than any one model's
weakness, so it is a standard control rather than a hand-optimized prompt.
(iii) \emph{text-tools}: \texttt{tools=} dropped, the same text guidance in the
prompt, calls parsed from text; uniform across all models. We additionally probe a
constrained-decoding setup (\S\ref{sec:results}).
\paragraph{Models.} Qwen2.5-Coder 0.5B/1.5B/3B/7B/14B, Llama-3.2-3B, Phi-3-mini,
Gemma-3-4B, Gemma-3-270m, served locally via Ollama, plus \texttt{deepseek-v4-flash}
(cloud) as a high-capability anchor. The local set spans the checkpoints a
practitioner actually obtains from a default \texttt{ollama pull} at laptop scale and
deliberately mixes models with and without native tool training, because that mix is
what exposes the per-model gate; it is a stress test of the serving layer rather than
a survey of current agent-oriented models, and the cloud anchor is considerably newer
than the local set. Exact tags, quantization, and versions are in \S\ref{sec:repro}.

\section{Results}
\label{sec:results}
\begin{figure*}[t]
\centering
\includegraphics[width=0.92\textwidth]{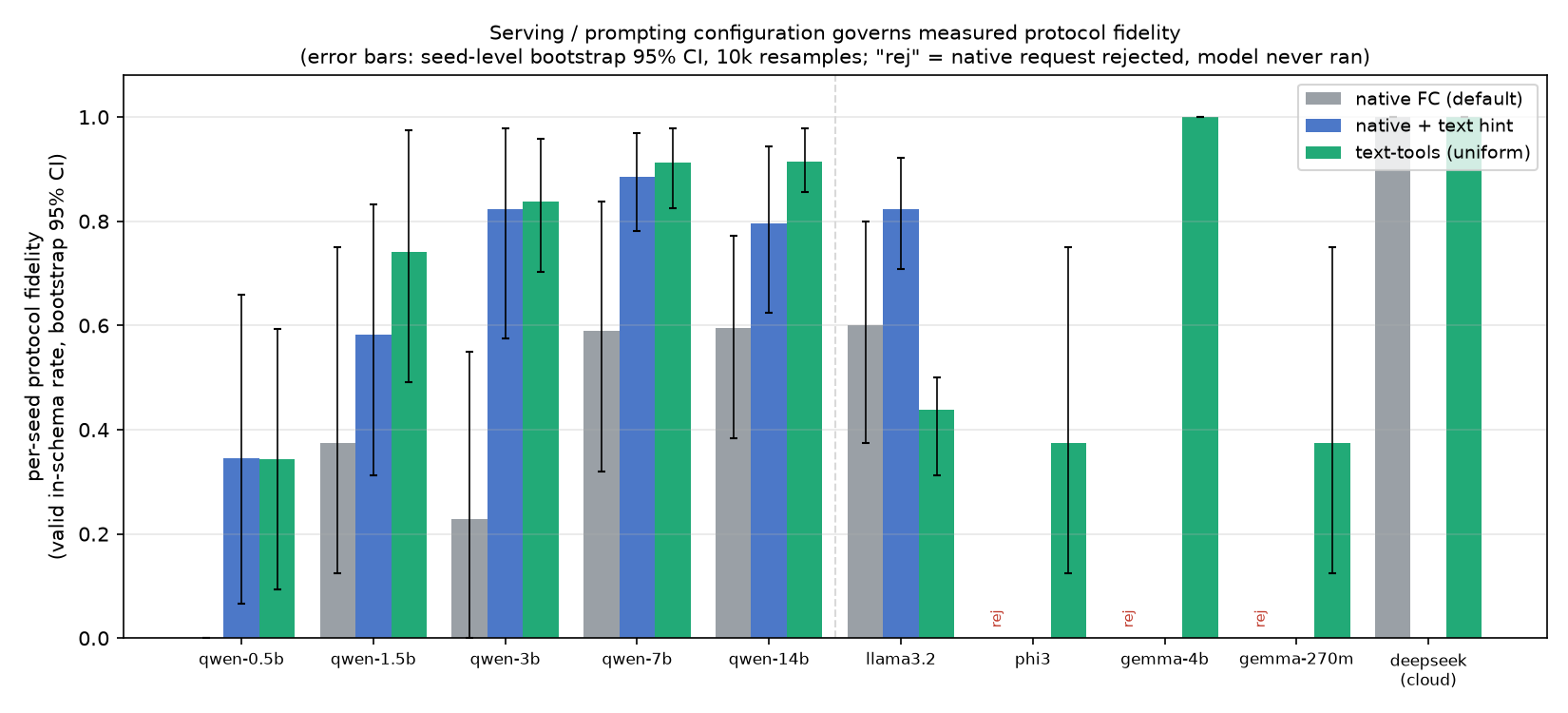}
\caption{Per-seed protocol fidelity (valid in-schema rate, seed-level bootstrap
$95\%$ CI over 8 seeds, $10{,}000$ resamples) under three configurations. ``rej'' marks models whose native request was
rejected by Ollama on every seed (the model never ran). For accepted models, adding
a text hint to the native call (blue) recovers most of the fidelity, so the native
channel itself is not the bottleneck; Llama-3.2 is the exception, where the uniform
text-tools protocol (green) discards its real native support and lowers fidelity.}
\label{fig:modes}
\end{figure*}

\begin{figure*}[t]
\centering
\includegraphics[width=0.82\textwidth]{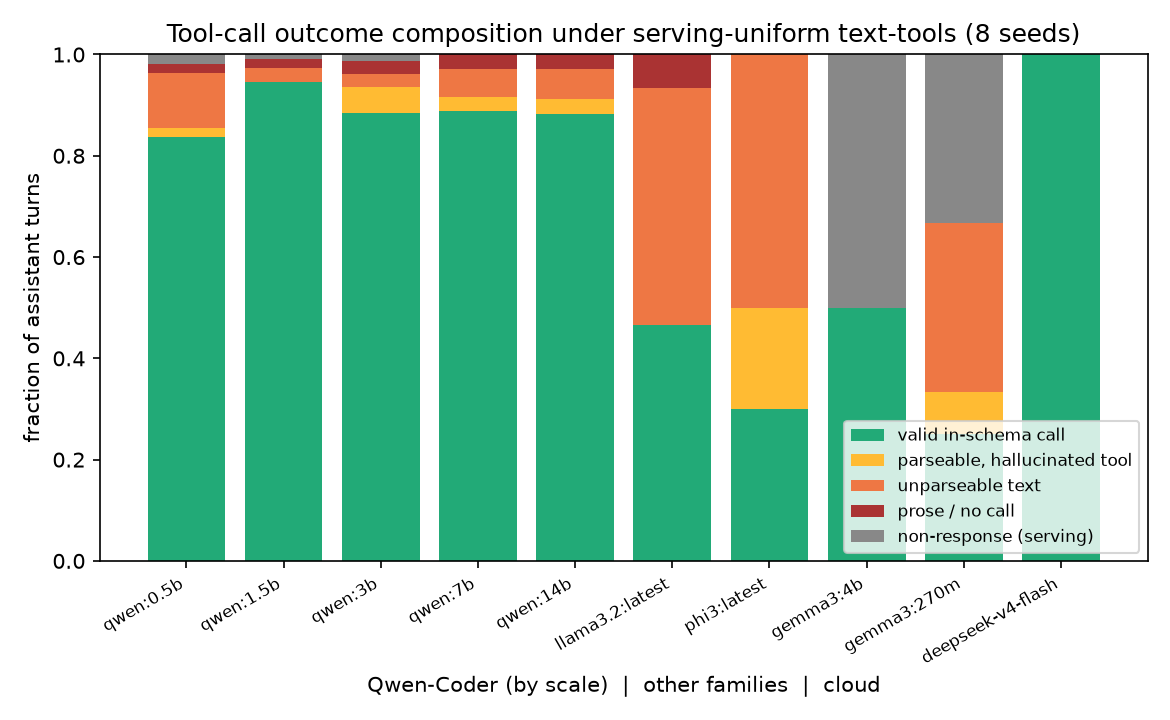}
\caption{Per-turn outcome composition under the uniform text-tools protocol (8
seeds), including a non-response band. Gemma-3-4B's apparent all-valid behavior is
half non-responses; the fidelity rate (valid over produced turns) excludes these,
which is why it must be reported alongside the non-response share, not alone.}
\label{fig:comp}
\end{figure*}

\paragraph{Native FC mismeasures, two ways
(Fig.~\ref{fig:modes}, gray; Fig.~\ref{fig:comp}).} Under native FC, a naive
per-turn analysis reports Phi-3 and Gemma-3 at $0\%$; in fact $100\%$ of their turns
are rejected requests, so the model never ran and the rate is properly undefined
(marked ``rej'' in Fig.~\ref{fig:modes} and Table~\ref{tab:main}). The second mechanism is non-response:
Gemma-3-4B, for instance, emits one valid call per seed and then fails to respond,
which the naive denominator would count against it (Fig.~\ref{fig:comp}, gray band).
\paragraph{The prompt, not the channel, drives the accepted-model gap
(Fig.~\ref{fig:modes}).} For every model the server accepts, adding the text hint
while keeping the native channel raises per-seed fidelity substantially (Qwen
0.5B--14B: $0/38/23/59/60\%$ native $\to 35/58/82/89/80\%$ native+hint), and
native+hint is close to the uniform text-tools rate. So the low native numbers
reflect a default call path without explicit format guidance, not model inability.
Llama-3.2 is the informative exception: native+hint reaches $82\%$ but the uniform
text-tools protocol drops it to $44\%$, because Llama-3.2 has real native
\texttt{tool\_calls} support that the text protocol discards. The best-performing interface
is model-dependent, so no single serving configuration maximizes measured fidelity for
every model.
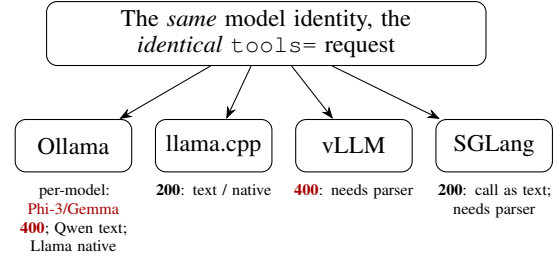
\begin{figure}[t]
\centering
\begin{tikzpicture}[
  >=Stealth,
  stk/.style={draw,rounded corners,align=center,text width=1.4cm,
              inner sep=2pt,minimum height=1.8em,font=\footnotesize},
  res/.style={align=center,text width=1.6cm,font=\tiny,inner sep=1.5pt}]
\node[draw,rounded corners,align=center,text width=5.6cm,inner sep=3pt,font=\small]
  (req) {The \emph{same} model identity, the \emph{identical} \texttt{tools=} request};
\node[stk,below=7mm of req,xshift=-2.55cm] (ol) {Ollama};
\node[stk,right=3mm of ol] (lc) {llama.cpp};
\node[stk,right=3mm of lc] (vl) {vLLM};
\node[stk,right=3mm of vl] (sg) {SGLang};
\draw[->] (req)--(ol); \draw[->] (req)--(lc); \draw[->] (req)--(vl); \draw[->] (req)--(sg);
\node[res,below=1mm of ol] {per-model: \textcolor{red!65!black}{Phi-3/Gemma \textbf{400}};
  Qwen text; Llama native};
\node[res,below=1mm of lc] {\textbf{200}: text / native};
\node[res,below=1mm of vl] {\textcolor{red!65!black}{\textbf{400}}: needs parser};
\node[res,below=1mm of sg] {\textbf{200}: call as text; needs parser};
\end{tikzpicture}
\caption{\textbf{Identical request, different serving outcomes.} Four serving stacks
handle the corresponding model checkpoints and identical \texttt{tools=} request
differently, so what a naive
evaluator records as model (in)ability is set by the serving stack and its
configuration, not the model. (Ollama and llama.cpp probed across models; vLLM and
SGLang default behavior confirmed on Qwen/Phi-3.)}
\label{fig:threestacks}
\end{figure}

\paragraph{The rejection is stack policy, not the model
(Fig.~\ref{fig:threestacks}).} The same GGUF weights that
Ollama rejects with HTTP~400 run when served by llama.cpp: Phi-3 and Gemma-3 both
return a (text) response under an identical \texttt{tools=} request, while
Qwen-0.5B's text call and Llama-3.2's native call are reproduced on both stacks
(Table~\ref{tab:cross}). These apparent $0\%$ outcomes are therefore artifacts of
that stack's tool-gating policy rather than measurements of model behavior. A third stack,
vLLM, makes the point sharper by failing differently: with its default launch it
refuses the same \texttt{tools=} request outright (HTTP~400, ``\texttt{'auto' tool
choice requires -{}-enable-auto-tool-choice and -{}-tool-call-parser}''), independent
of the model: the identical error is returned for the models we probed in the
default vLLM path (Qwen-0.5B and Phi-3), as the check precedes model dispatch. Enabling those flags with the
\texttt{hermes} parser makes vLLM accept
the request (HTTP~200), but for the two small models we probed the call still did not
surface as a native \texttt{tool\_call}: Qwen2.5-Coder-0.5B emitted the call as a
fenced JSON block the parser did not extract, and Phi-3 produced prose. A fourth
stack, SGLang, differs from vLLM again: by default it accepts the request (HTTP~200)
but returns the call as text rather than a native \texttt{tool\_call} unless launched
with \texttt{-{}-tool-call-parser} (confirmed on Qwen-0.5B and Phi-3), so even the two
modern production stacks disagree on the default handling. Whether a
request is refused, and whether a call is recognized once accepted, are governed by
the serving stack and its launch configuration, not by the model.
\paragraph{The fidelity number is not robust (Table~\ref{tab:main}).} Turn-pooled
and per-seed rates diverge sharply when a model produces one long looping episode.
Qwen-0.5B under text-tools is $85\%$ pooled but $34\%$ per-seed: seven of eight
episodes fail in one or two turns while a single 41-turn episode of repeated valid
calls dominates the pool. Denominators are as small as 8--16 turns over 8 seeds for
the weaker models, so the seed-level bootstrap $95\%$ CIs are correspondingly wide:
Qwen-0.5B text-tools is $34\%$ $[9,59]$, Phi-3 $38\%$ $[12,75]$, and Gemma-3-270m
$38\%$ $[12,75]$ (Fig.~\ref{fig:modes} error bars), intervals far too wide to rank
these models against each other.
\paragraph{Constrained decoding removes the protocol failures, at a cost.} As a
baseline we constrain decoding to a JSON schema whose \texttt{name} field is an enum
of the available tools (Ollama structured outputs). On a single tool-call step, all
nine local models, including the three that scored lowest under text-tools
(Llama-3.2, Phi-3, Gemma-3-270m), emit a valid in-schema call on 8 of 8 trials:
parseability and in-schema names hold by construction. We do not run this as a full
agentic condition because it removes the model's ability to stop. Forced to emit a
tool call on every turn, the weaker models never terminate, producing 600--900-turn
loops within a single episode (Qwen-0.5B reached 869 assistant turns in one seed
before we cut it off). Constrained decoding thus fixes the protocol layer we measure
but trades an unparseable-call failure for a non-termination failure on weak models,
a further sign that the configuration governs the observed failure.
\paragraph{Replication on a second task (Table~\ref{tab:chain}).} Because Ollama
refuses the \texttt{tools=} request before the prompt is processed, the rejection is a
property of the model--serving pair and should be task-independent. We confirm this on
a structurally different dependency-chain task (trace a multi-hop chain across modules
and implement \texttt{compute\_total}, scored by import). The native tool-gating
replicates exactly: Phi-3 and Gemma-3 are rejected with HTTP~400, Llama-3.2 returns a
native call, and Qwen is accepted as text. Under the text-tools protocol (4 seeds),
the same qualitative pattern recurs: per-seed fidelity rises roughly with Qwen scale
($44\%\to100\%$) with the pooled rate again inflated by looping episodes; Gemma-3-4B
again emits a valid call and then fails to respond (4 non-responses); and the
magnitudes are task-dependent (Llama-3.2 falls to $12\%$ per-seed here, versus $60\%$
on aggregation) and noisy at these small denominators. The replication supports the
measurement caveat and the fragility of the numbers, not a stable cross-model ranking.
\paragraph{Replication on HumanEval.} To check the confound is not an artifact of our
own task design, we repeat the single-turn measurement on HumanEval
\cite{humaneval2021}, a widely used code benchmark, framing each problem as a tool-call
task in which the model submits its solution by calling one tool. The native gating
replicates exactly: Phi-3 and Gemma-3 are rejected with HTTP~400 while Qwen and
Llama-3.2 are accepted. Under the text-tools protocol the valid-call rate again rises
with Qwen scale ($4/6$, $5/6$, $6/6$ for 0.5B/1.5B/3B) and the family-specific failure
modes recur (Gemma-3-270m $0/6$, every response unparseable prose; Gemma-3-4B $2/6$),
while Llama-3.2 reaches $6/6$. Magnitudes are task-dependent and $n{=}6$, so this is an
external-validity check on the gating and the failure modes, not a precise rate.

\paragraph{What we do not claim.} The per-seed rates show some variation with model
size within the Qwen family and across families, but the intervals overlap and the
estimates depend on the pooling choice, so we do not claim a scale law, a family or
tool-training effect, or a dissociation from reasoning. We ran a small no-tools
reasoning probe while exploring those questions; it was not conclusive and we omit
it to avoid over-reading.

\begin{table}[t]
\centering
\small
\setlength{\tabcolsep}{4pt}
\begin{tabular}{lccc}
\toprule
model & native & native+hint & text-tools \\
 & seed (pool) & seed (pool) & seed (pool) \\
\midrule
Qwen-0.5B  & $0\,(0)$    & $35\,(29)$ & $34\,(85)$ \\
Qwen-1.5B  & $38\,(24)$  & $58\,(91)$ & $74\,(95)$ \\
Qwen-3B    & $23\,(79)$  & $82\,(96)$ & $84\,(90)$ \\
Qwen-7B    & $59\,(79)$  & $89\,(80)$ & $91\,(89)$ \\
Qwen-14B   & $60\,(74)$  & $80\,(71)$ & $92\,(88)$ \\
\midrule
Llama-3.2-3B & $60\,(62)$ & $82\,(89)$ & $44\,(47)$ \\
Phi-3-mini   & rej        & ---        & $38\,(30)$ \\
Gemma-3-4B   & rej        & ---        & $100\,(100)$ \\
Gemma-3-270m & rej        & ---        & $38\,(38)$ \\
\midrule
deepseek (cloud) & $100\,(100)$ & --- & $100\,(100)$ \\
\bottomrule
\end{tabular}
\caption{Protocol fidelity as per-seed mean with turn-pooled rate in parentheses,
in percent. ``rej'' = the native request was rejected on all 8 seeds, so native
fidelity is undefined, not $0\%$. Per-seed and pooled diverge most where a model
loops (e.g.\ Qwen-0.5B text-tools $34$ vs $85$), which is why we report both.}
\label{tab:main}
\end{table}

\begin{table}[t]
\centering
\footnotesize
\setlength{\tabcolsep}{3pt}
\begin{tabular}{lcccc}
\toprule
model & Ollama & llama.cpp & vLLM & SGLang \\
\midrule
Qwen-0.5B    & 200 txt & 200 txt & 400$^\dagger$ & 200 txt \\
Llama-3.2    & 200 nat & 200 nat & ---           & --- \\
Phi-3        & 400 rej & 200 txt & 400$^\dagger$ & 200 txt \\
Gemma-3-270m & 400 rej & 200 txt & ---           & --- \\
\bottomrule
\end{tabular}
\caption{Cross-stack handling of the same \texttt{tools=} request for the same models
(identical GGUF weights on Ollama and llama.cpp; upstream Hugging Face checkpoints on
the vLLM and SGLang probes). Cells give the HTTP status and mode:
\texttt{txt}\,=\,call returned as text, \texttt{nat}\,=\,native \texttt{tool\_call},
\texttt{rej}\,=\,request rejected. Models Ollama rejects per model (the silent $0\%$) run on
llama.cpp, and models both stacks serve are handled consistently.
$^\dagger$vLLM refuses the request \emph{by default}, independent of the model, until
launched with \texttt{-{}-enable-auto-tool-choice} and a \texttt{-{}-tool-call-parser}
(once enabled, neither probed model produced a native call). SGLang instead accepts the
request by default but returns an unparsed text call unless launched with
\texttt{-{}-tool-call-parser}. vLLM and SGLang were confirmed on Qwen-0.5B and Phi-3;
\mbox{---}\,=\,not probed on that stack. The handling is a stack-and-configuration
policy, not a model property.}
\label{tab:cross}
\end{table}

\begin{table}[t]
\centering
\small
\begin{tabular}{lcc}
\toprule
model & native FC & text-tools seed (pool), $n$ \\
\midrule
Qwen-0.5B  & text       & $44\,(80)$, 15 \\
Qwen-1.5B  & text       & $81\,(88)$, 16 \\
Qwen-3B    & text       & $87\,(83)$, 12 \\
Qwen-7B    & text       & $85\,(88)$, 26 \\
Qwen-14B   & text       & $100\,(100)$, 29 \\
Llama-3.2-3B & native   & $12\,(20)$, 5 \\
Phi-3-mini   & rejected  & $50\,(50)$, 4 \\
Gemma-3-4B   & rejected  & $100\,(100)$, 4 \\
Gemma-3-270m & rejected  & $0\,(0)$, 4 \\
\bottomrule
\end{tabular}
\caption{Second task (dependency-chain), 4 seeds. ``native FC'' is the serving
handling (identical to Table~\ref{tab:main}: Qwen accepted as text, Llama native,
Phi-3/Gemma rejected with HTTP~400). ``text-tools'' is per-seed fidelity with the
pooled rate in parentheses and $n$ produced turns. The native gating, the
pooled-vs-per-seed gap, and Gemma-3-4B's valid-then-non-respond pattern all recur;
magnitudes are task-dependent and noisy at these small $n$. Gemma-3-4B's $100\%$ comes
from only 4 produced turns amid non-responses and should be read together with
Fig.~\ref{fig:comp}, not as strong performance.}
\label{tab:chain}
\end{table}

\section{The harness conflates serving failures with model non-calls}
\label{sec:audit}
The contamination is silent because of how the trajectory represents the failure, not
because the failure leaves no trace. The transport evidence does arrive: Ollama
refuses the request with HTTP~400 and the message ``does not support tools''. But in
the stack we used, when the request is rejected or retries are exhausted, the
inference wrapper collapses that response into an assistant message whose content is a
generic error string (e.g.\ \texttt{"Failed to get response after multiple retries."}),
and the error type is not persisted into the saved trajectory. \textbf{The transport
failure is therefore detectable, but its structured failure type is not preserved.} A
downstream per-turn analysis that reads the message stream sees an ordinary non-call
assistant turn and attributes it to the model, unless it knows to string-match the
harness's specific error text; we separated rejection and non-response from genuine
model behavior only by adding such a match. Structured benchmarks like BFCL
\cite{bfcl2025} do track decode and empty-response states, and harnesses that keep
structured per-step records can too: what fails is the particular combination we
instrument, in which only the message stream is retained. This is a correctable defect
in the trajectory representation, and preserving structured error metadata is the first
item on our checklist (\S\ref{sec:discussion}). The per-model tool-gating that produces
the rejections in the first place is a property of the serving stack rather than the
benchmark, and better logging does not remove it.

\section{Reproducibility}
\label{sec:repro}
The primary Ollama and llama.cpp runs are local and CPU-only, except for the
cloud anchor. The vLLM and SGLang cross-stack probes use standard GPU serving
setups. Models are Ollama tags
\texttt{qwen2.5-coder:\{0.5b,\allowbreak 1.5b,\allowbreak 3b,\allowbreak 7b,\allowbreak 14b\}}, \texttt{llama3.2:latest},
\texttt{phi3:latest}, \texttt{gemma3:\{4b,270m\}} at Ollama's default quantization
(Q4\_K\_M for these tags); the cloud anchor is \texttt{deepseek-v4-flash}. The
serving behavior is Ollama-version-dependent: the per-model \texttt{tools=} gating
and the HTTP~400 ``does not support tools'' contract are properties of the Ollama
release (we used Ollama 0.30.8) and must be pinned to reproduce the native-mode
result. Our native-mode claims are scoped to that release: the mechanism (a per-model
template flag consulted before dispatch) is structural, but which tags are gated is a
release-level policy that can change, and we do not claim the specific per-model
outcomes hold for later releases. The
cross-stack check uses llama.cpp \texttt{llama-server} with \texttt{-{}-jinja} on the
same GGUF weights. The vLLM and SGLang probes use Qwen-0.5B and Phi-3 on the upstream
Hugging Face checkpoints on a GPU host (vLLM on a T4, SGLang on an A800~80GB): by default
vLLM rejects the \texttt{tools=} request until launched with
\texttt{-{}-enable-auto-tool-choice} and a \texttt{-{}-tool-call-parser}, while SGLang
accepts it but returns the call as text unless launched with
\texttt{-{}-tool-call-parser}; the exact package versions and launch commands are in
the released probe scripts. Decoding used temperature $1.0$, top-$p$ $1.0$. The three
conditions correspond to environment flags \texttt{(none)}, \texttt{TOOL\_HINT}, and
\texttt{TEXT\_TOOLS}; the patch to the harness, the task config, the per-turn
classifier, the bootstrap-CI script, the figures, and the probes are available at \url{https://github.com/LijuanTang94/serving-confound-repo}.
\textbf{Scope of reproducibility.} The qualitative findings are deterministic and
version-pinned: the per-model rejection, the cross-stack contrast, and the
prompt-recovers-fidelity result reproduce exactly given the same Ollama release.
The per-seed magnitudes do not reproduce to the digit, because decoding is sampled
($T{=}1.0$) and our seeds index task instances rather than the sampling RNG; a
re-run yields the same pattern (rejection, looping-inflated pooled rates, thin-$n$
variance) with different exact percentages. This is the fragility we report, not a
defect of it.

\section{Discussion: why this will get worse}
\label{sec:discussion}
The confound we report is not a one-off quirk of one library. It is a structural
consequence of how local agent stacks are assembled, and the trend is toward more
layers between the harness and the model, not fewer. A modern agent request passes
through a harness, a tool-protocol adapter, a serving engine (Ollama, vLLM,
llama.cpp), and a per-model chat template, each of which can accept, rewrite, or
reject a tool call independently of the model's ability. The Model Context Protocol
and similar tool-calling standards plausibly add yet another translation step, though
we have not measured this and offer it as conjecture rather than a finding. Hosted
APIs are not exempt in principle---a provider that refuses a \texttt{tools=} request
for an unsupported model, or returns a call the client fails to parse, yields the same
recorded outcome---but that gate is opaque and we did not probe it. Every such
layer is a place where a request can fail for reasons that have nothing to do with
the model, and where that failure can be logged in a way a naive evaluator reads as
a model error. As more practitioners evaluate small or local agents for cost and
privacy reasons, and as the stacks they use grow more complex, the gap between
``what the model can do'' and ``what the measurement records'' widens.

\paragraph{What is the evaluation target?} A fixed serving interface is appropriate
when the goal is to compare model behavior under a standardized protocol: the
interface is then part of the controlled measurement instrument. A different target is
the performance of a deployable model--server system, where model-specific parsers,
templates, or serving options may reasonably be enabled. These two questions should not
be conflated. Our checklist recommendation to hold the serving interface fixed applies
to standardized model comparisons; system-level evaluations should instead report the
full model--server configuration as part of the system being evaluated.

\noindent The remedy is cheap but has to be deliberate:

\begin{center}
\fbox{\begin{minipage}{0.93\columnwidth}
\textbf{A protocol for measuring local-agent tool use.}\\[2pt]
\emph{Measure:}
\begin{enumerate}\setlength{\itemsep}{1pt}\setlength{\parskip}{0pt}\setlength{\topsep}{1pt}\setlength{\leftmargin}{1em}
  \item For standardized model comparisons, hold the serving interface fixed across
  models and pin and report the stack and its version.
  \item Log a transport- or serving-level failure (rejected request, timeout, empty
  response) as a \emph{distinct outcome}, never as a model non-call.
  \item Report per-seed rates with intervals, not a single turn-pooled number.
\end{enumerate}
\emph{Diagnose a $0\%$ tool-call rate, in order:}
\begin{enumerate}\setlength{\itemsep}{1pt}\setlength{\parskip}{0pt}\setlength{\topsep}{1pt}\setlength{\leftmargin}{1em}
  \item Did the serving layer refuse or empty the request (HTTP~4xx, retry
  exhaustion)? If so, the model never ran.
  \item Is a tool-call parser configured for this stack and model? An accepted request
  can still surface the call as text.
  \item Only after (1)--(2) are ruled out, attribute the failure to the model.
\end{enumerate}
\end{minipage}}
\end{center}

\noindent A refused or empty request is thus a first-class evaluation outcome, not a
framework-compatibility bug: any benchmark or harness that runs models through a
serving stack should record and report it as its own category. Concretely, future
local-agent benchmarks should report the serving configuration (stack, version,
tool-call parser) alongside the model identity, just as they already report decoding
parameters and hardware. Treating this as a
measurement-design requirement, rather than a per-tool quirk, is becoming as important
to agent evaluation as the benchmarks themselves.

\section{Conclusion}
Native function calling on a local server gates the tool-call request per model and
records a refused or empty request as an ordinary non-call turn, so a capable model
can be measured at $0\%$ without ever running. A control arm shows that for accepted
models the default native path under-measures fidelity for lack of prompt guidance
rather than ability, that forcing a uniform text protocol can instead hurt a model
with real native support, and a cross-stack check confirms the rejection is stack
policy. Together with the large gap between pooled and per-seed estimates, the lesson
is a measurement checklist for small or local agents: for standardized model
comparisons hold the serving interface fixed, separate a refused or empty request
from a model non-call, and report per-seed rates with intervals. Otherwise the instrument measures the serving stack,
not the model.

\section*{Limitations}
\textbf{Serving stacks and scope.} The confound is not specific to one stack: we
confirm the per-request gating on \emph{four} serving stacks (Ollama, llama.cpp, vLLM,
SGLang; Table~\ref{tab:cross}), which handle the identical request differently. We
report per-seed fidelity on two tasks (aggregation, Table~\ref{tab:main}, 8 seeds;
dependency-chain, Table~\ref{tab:chain}, 4 seeds), both on Ollama, plus a single-turn
replication on HumanEval; the gating and the fragility of the numbers replicate across
all three tasks, but per-seed magnitudes are task-dependent, so we do not generalize
them. Whether the per-seed fidelity magnitudes transfer across stacks is untested;
request handling is stack- and
configuration-dependent across the four systems we probe. The vLLM and SGLang probes cover only
Qwen-0.5B and Phi-3, so their default-handling results are claims about those two
model--stack pairs rather than about either stack in general. \textbf{Thin denominators and
pooling.} The weaker models produce 8--16 turns over 8 seeds; per-seed and pooled
rates diverge, and we report both. \textbf{Constrained-decoding
baseline is single-turn.} Its numbers come from a single tool-call step, not a full
agentic run, because the agentic version did not terminate on weak models; it shows
parseability is recoverable but not multi-turn behavior under constraint.
\textbf{Hint design.} The text hint includes an allowed-name list; a weaker hint
might recover less, so ``the prompt drives it'' is specific to this guidance.
\textbf{Omitted probe.} We collected a reasoning probe but omit it because it was
inconclusive; we therefore make no reasoning claim. \textbf{Few-seed fragility.} An
earlier 3-seed pilot of ours showed a clean curve that 8 seeds dissolved;
small-model rates are high-variance.

\bibliography{refs}

\begin{thebibliography}{14}
\providecommand{\natexlab}[1]{#1}

\bibitem[{Cemri et~al.(2025)Cemri, Pan, Yang, Agrawal, Chopra, Tiwari, Keutzer,
  Parameswaran, Klein, Ramchandran, Zaharia, Gonzalez, and Stoica}]{mast2025}
Mert Cemri, Melissa~Z. Pan, Shuyi Yang, Lakshya~A. Agrawal, Bhavya Chopra,
  Rishabh Tiwari, Kurt Keutzer, Aditya Parameswaran, Dan Klein, Kannan
  Ramchandran, Matei Zaharia, Joseph~E. Gonzalez, and Ion Stoica. 2025.
\newblock \href {https://arxiv.org/abs/2503.13657} {Why do multi-agent {LLM}
  systems fail?}
\newblock \emph{arXiv preprint arXiv:2503.13657}.

\bibitem[{Chen et~al.(2021)}]{humaneval2021}
Mark Chen et~al. 2021.
\newblock Evaluating large language models trained on code.
\newblock \emph{arXiv preprint arXiv:2107.03374}.

\bibitem[{Lee et~al.(2025)Lee, Song, Han, Pyun, and Jo}]{patool2025}
Jonggeun Lee, Woojung Song, Jongwook Han, Haesung Pyun, and Yohan Jo. 2025.
\newblock \href {https://arxiv.org/abs/2510.07248} {Don't adapt small language
  models for tools; adapt tool schemas to the models}.
\newblock \emph{arXiv preprint arXiv:2510.07248}.
\newblock ACL 2026.

\bibitem[{Liu et~al.(2024)}]{agentbench2024}
Xiao Liu et~al. 2024.
\newblock {AgentBench}: Evaluating {LLMs} as agents.
\newblock In \emph{International Conference on Learning Representations
  (ICLR)}.

\bibitem[{Lu et~al.(2024)}]{toolsandbox2024}
Jiarui Lu et~al. 2024.
\newblock {ToolSandbox}: A stateful, conversational, interactive evaluation
  benchmark for {LLM} tool use capabilities.
\newblock \emph{arXiv preprint arXiv:2408.04682}.

\bibitem[{Pape et~al.(2026)Pape, Evertz, and Sch{\"o}nherr}]{silenthyper2026}
David Pape, Jonathan Evertz, and Lea Sch{\"o}nherr. 2026.
\newblock \href {https://arxiv.org/abs/2605.19537} {The silent hyperparameter:
  Quantifying the impact of inference backends on {LLM} reproducibility}.
\newblock \emph{arXiv preprint arXiv:2605.19537}.

\bibitem[{Patil et~al.(2025)Patil, Mao et~al.}]{bfcl2025}
Shishir~G. Patil, Huanzhi Mao, et~al. 2025.
\newblock The {Berkeley} function calling leaderboard ({BFCL}): From tool use
  to agentic evaluation of large language models.
\newblock In \emph{International Conference on Machine Learning (ICML)}.
\newblock Leaderboard: \url{https://gorilla.cs.berkeley.edu/leaderboard.html}.

\bibitem[{Patil et~al.(2023)Patil, Zhang, Wang, and Gonzalez}]{gorilla2023}
Shishir~G. Patil, Tianjun Zhang, Xin Wang, and Joseph~E. Gonzalez. 2023.
\newblock {Gorilla}: Large language model connected with massive {APIs}.
\newblock \emph{arXiv preprint arXiv:2305.15334}.

\bibitem[{Willard and Louf(2023)}]{outlines2023}
Brandon~T. Willard and R{\'e}mi Louf. 2023.
\newblock Efficient guided generation for large language models.
\newblock \emph{arXiv preprint arXiv:2307.09702}.

\bibitem[{Yang et~al.(2024)}]{sweagent2024}
John Yang et~al. 2024.
\newblock {SWE-agent}: Agent-computer interfaces enable automated software
  engineering.
\newblock \emph{arXiv preprint arXiv:2405.15793}.

\bibitem[{Yao et~al.(2024)}]{taubench2024}
Shunyu Yao et~al. 2024.
\newblock $\tau$-bench: A benchmark for tool-agent-user interaction in
  real-world domains.
\newblock \emph{arXiv preprint arXiv:2406.12045}.

\bibitem[{Yao et~al.(2026)Yao, Tan, Liu, Li, Wang, Yu, Tan, Tian, Zhao, Sun,
  Zhang, and Yang}]{harnessbench2026}
Yilun Yao, Xinyu Tan, Chao-Hsuan Liu, Yaoming Li, Zhengyang Wang, Wenhan Yu,
  Zhewen Tan, Yuxuan Tian, Guangxiang Zhao, Lin Sun, Xiangzheng Zhang, and Tong
  Yang. 2026.
\newblock \href {https://arxiv.org/abs/2605.27922} {{Harness-Bench}: Measuring
  harness effects across models in realistic agent workflows}.
\newblock \emph{arXiv preprint arXiv:2605.27922}.

\bibitem[{Zeng et~al.(2026)Zeng, Huang, and He}]{locabench2026}
Weihao Zeng, Yuzhen Huang, and Junxian He. 2026.
\newblock \href {https://arxiv.org/abs/2602.07962} {{LOCA-bench}: Benchmarking
  language agents under controllable and extreme context growth}.
\newblock \emph{arXiv preprint arXiv:2602.07962}.

\bibitem[{Zhang et~al.(2023)Zhang, Chen, Li, and Wang}]{tooldec2023}
Kexun Zhang, Hongqiao Chen, Lei Li, and William Wang. 2023.
\newblock \href {https://arxiv.org/abs/2310.07075} {Don't fine-tune, decode:
  Syntax error-free tool use via constrained decoding}.
\newblock \emph{arXiv preprint arXiv:2310.07075}.

\end{thebibliography}

\end{document}